\documentclass[10pt,twocolumn,letterpaper]{article}

\usepackage{titling}

\usepackage{cvpr}
\usepackage{times}
\usepackage{epsfig}
\usepackage{graphicx}
\usepackage{amsmath}
\usepackage{amssymb}
\usepackage{algpseudocode}
\usepackage{algorithm}

\usepackage{booktabs}
\usepackage{array}

\usepackage[lofdepth,lotdepth]{subfig}
\DeclareGraphicsExtensions{.pdf,.jpg,.png}

\usepackage[pagebackref=true,breaklinks=true,letterpaper=true,colorlinks,bookmarks=false]{hyperref}

\cvprfinalcopy 

\def\cvprPaperID{2393} 

\ifcvprfinal\fi
\begin{document}

\title{Iterative Instance Segmentation}

\author{Ke Li\\
UC Berkeley\\
{\tt\small ke.li@eecs.berkeley.edu}
\and
Bharath Hariharan\\
Facebook AI Research\\
{\tt\small bharathh@fb.com}
\and
Jitendra Malik\\
UC Berkeley\\
{\tt\small malik@eecs.berkeley.edu}
}

\maketitle
\thispagestyle{empty}

\begin{abstract}

Existing methods for pixel-wise labelling tasks generally disregard the underlying structure of labellings, often leading to predictions that are visually implausible. While incorporating structure into the model should improve prediction quality, doing so is challenging -- manually specifying the form of structural constraints may be impractical and inference often becomes intractable even if structural constraints are given. We sidestep this problem by reducing structured prediction to a sequence of unconstrained prediction problems and demonstrate that this approach is capable of automatically discovering priors on shape, contiguity of region predictions and smoothness of region contours from data without any a priori specification. On the instance segmentation task, this method outperforms the state-of-the-art, achieving a mean $\mathrm{AP}^{r}$ of $63.6\%$ at $50\%$ overlap and $43.3\%$ at $70\%$ overlap. 

\end{abstract}

\section{Introduction}

In computer vision, the objective of many tasks is to predict a pixel-wise labelling of the input image. While the intrinsic structure of images constrains the space of sensible labellings, existing approaches typically eschew leveraging such cues and instead predict the label for each pixel independently. Consequently, the resulting predictions may not be visually plausible. To mitigate this, a common strategy is to perform post-processing on the predictions using superpixel projections~\cite{hypercolumn} or conditional random fields (CRFs)~\cite{KrahenbuhlNIPS2011}, which ensures the final predictions are consistent with local appearance cues like colour and texture but fails to account for global object-level cues like shape. 

\begin{figure}
\centering
\includegraphics[width=0.5\textwidth]{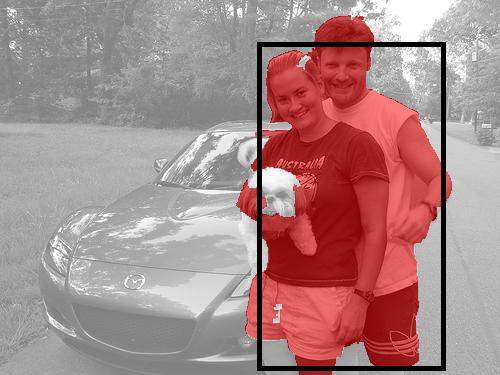}
\caption{A challenging image in which object instances are segmented incorrectly. While pixels belonging to the category are identified correctly, they are not correctly separated into instances.}
\label{fig:intro}
\end{figure}

Despite its obvious shortcomings, this strategy enjoys popularity, partly because incorporating global cues requires introducing higher-order potentials in the graphical model and often makes inference intractable. Because inference in general graphical models is NP-hard, extensive work on structured prediction has focused on devising efficient inference algorithms in special cases where the higher-order potentials take on a particular form. Unfortunately, this restricts the expressive power of the model. As a result, care must be taken to formulate the cues of interest as higher-order potentials of the desired form, which may not be possible. Moreover, low-energy configurations of the potentials often need to be specified manually a priori, which may not be practical when the cues of interest are complex and abstract concepts like shape. 

In this paper, we devise a method that learns implicit shape priors and use them to improve the quality of the predicted pixel-wise labelling. Instead of attempting to capture shape using explicit constraints, we would like to model shape implicitly and allow the concept of shape to emerge from data automatically. To this end, we draw inspiration from iterative approaches like auto-context~\cite{auto}, inference machines~\cite{RossCVPR2011} and iterative error feedback (IEF)~\cite{ief}. Rather than learning a model to predict the target in one step, we decompose the prediction process into multiple steps and allow the model to make mistakes in intermediate steps as long as it is able to correct them in subsequent steps. By learning to correct previous mistakes, the model must learn the underlying structure in the output implicitly in order to use it to make corrections. 

To evaluate if the method is successful in learning shape constraints, a perfect testbed is the task of instance segmentation, the goal of which is to identify the pixels that belong to each individual object instance in an image. Because the unit of interest is an object instance rather than an entire object category, methods that leverage only local cues have difficulty in identifying the instance a pixel belongs to in scenes with multiple object instances of the same category that are adjacent to one another, as illustrated in Figure~\ref{fig:intro}. We demonstrate that the proposed method is able to successfully learn a category-specific shape prior and correctly suppresses pixels belonging to other instances. It is also able to automatically discover a prior favouring contiguity of region predictions and smoothness of region contours despite these being not explicitly specified in the model. Quantitatively, it outperforms the state-of-the-art and achieves a mean $\mathrm{AP}^{r}$ of $63.6\%$ at $50\%$ overlap and $43.3\%$ at $70\%$ overlap. 

\section{Related Work}

Yang et al.~\cite{YangTPAMI2012} first described the task of segmenting out individual instances of a category. The metrics we use in this paper were detailed by Tighe et al.~\cite{TigheCVPR2014}, who proposed non-parametric transfer of instance masks from the training set to detected objects, and by Hariharan et al.~\cite{BharathECCV2014} who used convolutional neural nets (CNNs) ~\cite{LecunNC1989} to classify region proposals. We use the terminology and metrics proposed by the latter in this paper. Dai et al.~\cite{DaiCVPR2015} used ideas from~\cite{HeECCV2014} to speed up the CNN-based proposal classification significantly.

A simple way of tackling this task is to run an object detector and segment out each detected instance. The notion of segmenting out detected objects has a long history in computer vision. Usually this idea has been used to aid semantic segmentation, or the task of labeling pixels in an image with category labels. Borenstein and Ullman \cite{borenstein2002class} first suggested using category-specific information to improve the accuracy of segmentation. Yang et al.~\cite{YangTPAMI2012} start from object detections from the deformable parts model~\cite{FelzenszwalbPAMI2010} and paste figure-ground masks for each detected object. Similarly, Brox et al.~\cite{BroxCVPR2011} and Arbel\'{a}ez et al.~\cite{ArbelaezCVPR2012} paste figure-ground masks for poselet detections~\cite{BourdevECCV2010}. Recent advances in computer vision have all but replaced early detectors such as DPM and poselets with ones based on CNNs~\cite{LecunNC1989,GirshickCVPR2014,GirshickICCV2015} and produced dramatic improvements in performance in the process. In the CNN era, Hariharan et al.~\cite{hypercolumn} used features from CNNs to segment out R-CNN detections~\cite{GirshickCVPR2014}. 

When producing figure-ground masks for detections, most of these approaches predict every pixel independently. However, this disregards the fact that pixels in the image are hardly independent of each other, and a figure-ground labeling has to satisfy certain constraints. Some of these constraints can be simply encoded as local smoothness: nearby pixels of similar color should be labeled similarly. This can be achieved simply by aligning the predicted segmentation to image contours~\cite{BroxCVPR2011} or projecting to superpixels~\cite{hypercolumn}. More sophisticated approaches model the problem using CRFs with unary and pairwise potentials ~\cite{RotherTOG2004, ParkhiICCV2011, KrahenbuhlNIPS2011}. Later work considers extending these models by incorporating higher-order potentials of specific forms for which inference is tractable~\cite{kohli2013principled, li2013exploring}. A related line of work explores learning a generative model of masks~\cite{eslami2014shape} using a deep Boltzmann machine~\cite{salakhutdinov2009deep}. Zheng et al.~\cite{ZhengArxiv2015} show that inference in CRFs can be viewed as recurrent neural nets and trained together with a CNN to label pixels, resulting in large gains. Another alternative is to use eigenvectors obtained from normalized cuts as an embedding for pixels~\cite{MajiCVPR2011, MaireICCV2011}.

However, images contain more structure than just local appearance-dependent smoothness. For instance, one high informative form of global cue is shape; in the case of persons, it encodes important constraints like two heads cannot be part of the same person, the head must be above the torso and so on. There has been prior work on handling such constraints in the pose estimation task by using graphical models defined over keypoint locations~\cite{YangTPAMI2013, TompsonNIPS2014}. However, in many applications, keypoint locations are unknown and such constraints must be enforced on raw pixels. Explicitly specifying these constraints on pixels is impractical, since it would require formulating potentials that are capable of localizing different parts of an object, which itself is a challenging task. Even if this could be done, the potentials that are induced would be higher order (which arises from the relative position constraints among multiple parts of an object) and non-submodular (due to mutual exclusivity constraints between pixels belonging to two different heads). This makes exact inference and training in these graphical models intractable. 

\begin{figure*}[t]
    \centering
    \includegraphics[width=1.0\textwidth]{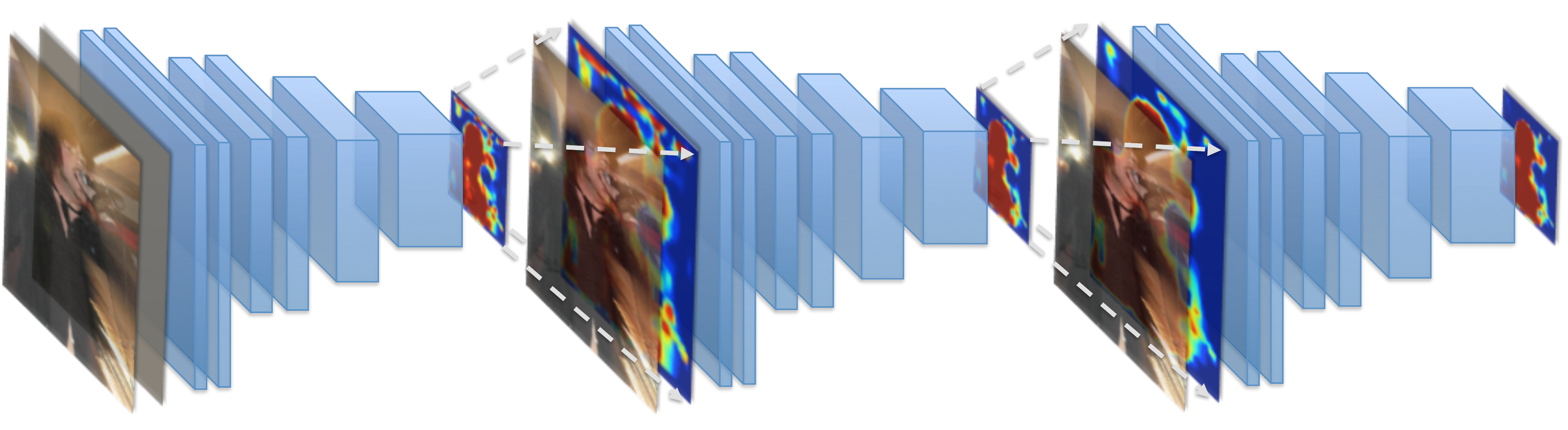}
    \caption{The proposed method decomposes the prediction process into multiple steps, each of which consists of performing unconstrained inference on the input image and the prediction from the preceding step. The diagram above illustrates a three-step prediction procedure when a convolutional neural net is used as the underlying model, as is the case with our method when applied to instance segmentation. }
    \label{fig:teaser}
\end{figure*}

Auto-context~\cite{auto} and inference machines~\cite{RossCVPR2011} take advantage of the observation that performing accurate inference does not necessarily require modelling the posterior distribution explicitly. Instead, these approaches devise efficient iterative inference procedures that directly approximate message passing. By doing so, they are able to leverage information from distant spatial locations when making predictions while remaining computationally efficient. Later work~\cite{rhinehart2015visual} extends this idea and derives an iterative prediction procedure that approximates the sequence of outputs of an oracle that has access to ground truth. In a similar spirit, other methods model the iterative process as recurrent neural nets ~\cite{PinheiroICML2014,ZhengArxiv2015}. IEF~\cite{ief} uses a related approach on the task of human pose estimation by directly refining the prediction rather than approximating message passing or oracle output in each iteration. While this approach shows promise when the predictions lie in a low-dimensional space of possible 2D locations of human joints, it is unclear if it will be effective when the output is high-dimensional and embeds complex structure like shape, as is the case with tasks that require a pixel-wise labelling of the input. In this paper, we devise an iterative method that supports prediction in high-dimensional spaces without a natural distance metric for measuring conformity to structure. 

\section{Method}

\subsection{Task and Setting}

The objective of the instance segmentation task, also known as simultaneous detection and segmentation (SDS), is to predict the segmentation mask for each object instance in an image. Typically, an object detection system is run in the first stage of the pipeline, which generates a set of candidate bounding boxes along with the associated detection scores and category labels. Next, non-maximum suppression (NMS) is applied to these detections, which are then fed into the segmentation system, which predicts a heatmap for each bounding box representing the probability of each pixel inside the bounding box belonging to the foreground object of interest. The heatmaps then optionally undergo some form of post-processing, such as projection to superpixels. Finally, they are binarized by applying a threshold, yielding the final segmentation mask predictions. We use fast R-CNN \cite{GirshickICCV2015} trained on MCG \cite{arbelaez2014multiscale} bounding box proposals as our detection system and focus on designing the segmentation system in this paper. 

\subsection{Segmentation System}

For our segmentation system, we use a CNN that takes a $224 \times 224$ patch as input and outputs a $50 \times 50$ heatmap prediction. The architecture is based on that of the hypercolmumn net proposed by Hariharan et al.~\cite{hypercolumn}, which is designed to be sensitive to image features at finer scales and relative locations of feature activations within the bounding box. Specifically, we use the architecture based on the VGG 16-layer net \cite{simonyan2014very} (referred to as ``O-Net" in~\cite{hypercolumn}), in which heatmaps are computed from the concatenation of upsampled feature maps from multiple intermediate layers, known as the hypercolumn representation. The CNN is trained end-to-end on the PASCAL VOC 2012 training set with ground truth instance segmentation masks from the Semantic Boundaries Dataset (SBD) \cite{sbd} starting from an initialization from the weights of a net finetuned for the detection task using R-CNN~\cite{GirshickCVPR2014}.

\subsection{Algorithm}

We would like to incorporate global cues like shape when making predictions. Shape encodes important structural constraints, such as the fact that a person cannot have two heads, which is why humans are capable of recognizing the category of an object from its silhouette almost effortlessly. So, leveraging shape enables us to disambiguate region hypotheses that all correctly cover pixels belonging to the category of interest but may group pixels into instances incorrectly. 

Producing a heatmap prediction that is consistent with shape cues is a structured prediction problem, with the structure being shape constraints. The proposed algorithm works by reducing the structured prediction problem to a sequence of unconstrained prediction problems. Instead of forcing the model to produce a prediction that is consistent with both the input and the structure in a single step, we allow the model to disregard structure initially and train it to correct its mistakes arising from disregarding structure over multiple steps, while ensuring consistency of the prediction with the input in each step. The final prediction is therefore consistent with both the input and the structure. Later, we demonstrate that this procedure is capable to learning a shape prior, a contiguity prior and a contour smoothness prior purely from data without any a priori specification to bias the learning towards finding these priors. 

At test time, in each step, we feed the input image and the prediction from the previous step, which defaults to constant prediction of $1/2$ in the initial step, into the model and take the prediction from the last step as our final prediction. In our setting, the model takes the form of a CNN. Please see Figure \ref{fig:teaser} for a conceptual illustration of this procedure. 

\begin{algorithm}
\footnotesize
\caption{Training Procedure}
\label{train_alg}
\begin{algorithmic}
\Require{$D$ is a training set consisting of $(x,y)$ pairs, where $x$ and $y$ denote the instance and the ground truth labelling respectively, and $f$ is the model}
\Function{Train}{$D$, $f$}
    \State \emph{// $p^{(t)}_{x}$ is the predicted labelling of $x$ in the $t^{\mathrm{th}}$ stage}
    \State $p^{(0)}_{x} \gets \left(\begin{array}{ccc}1/2 & \cdots & 1/2\end{array}\right)^{T} \; \forall \left(x,y\right) \in D$
    \For{$t =1$ \textbf{to} $N$}
        \State \emph{// Training set for the current stage}
        \State $T \gets \left\{ \left. \left(\left(\begin{array}{c}
x\\
p^{(i)}_x
\end{array}\right),y\right) \right| \left(x,y\right) \in D, i < t \right\}$ 
        \State Train model $f$ on $T$ starting from the current parameters of $f$
        \State $p^{(t)}_x \gets f\left(\begin{array}{c}
x\\
p^{(t-1)}_x
\end{array}\right) \; \forall \left(x,y\right) \in D$
    \EndFor
    \State \Return $f$
\EndFunction
\end{algorithmic}
\normalsize
\end{algorithm}
\vspace{-14pt}
\begin{algorithm}
\footnotesize
\caption{Testing Procedure}
\label{test_alg}
\begin{algorithmic}
\Require{$f$ is the model and $x$ is an instance}
\Function{Test}{$f$, $x$}
    \State \emph{// $\hat{y}^{(t)}$ is the predicted labelling of $x$ after $t$ iterations}
    \State $\hat{y}^{(0)} \gets \left(\begin{array}{ccc}1/2 & \cdots & 1/2\end{array}\right)^{T}$
    \For{$t = 1$ \textbf{to} $M$}
        \State $\hat{y}^{(t)} \gets f\left(\begin{array}{c}
x\\
\hat{y}^{(t-1)}
\end{array}\right)$
    \EndFor
    \State \Return $\hat{y}^{(M)}$
\EndFunction
\end{algorithmic}
\normalsize
\end{algorithm}

Training the model is straightforward and is done in stages: in the first stage, the model is trained to predict the ground truth segmentation mask with the previous heatmap prediction set to $1/2$ for all pixels and the predictions of the model at the end of training are stored for later use. In each subsequent stage, the model is trained starting from the parameter values at the end of the previous stage to predict the ground truth segmentation mask from the input image and a prediction for the image generated during any of the preceding stages. 

Pseudocode of the training and testing procedures are shown in Algorithms~\ref{train_alg} and \ref{test_alg}. 

\subsection{Discussion}

Modelling shape constraints using traditional structured prediction approaches would be challenging for three reasons. First, because the notion of shape is highly abstract, it is difficult to explicitly formulate the set of structural constraints it imposes on the output. Furthermore, even if it could be done, manual specification would introduce biases that favour human preconceptions and lead to inaccuracies in the predictions. Therefore, manually engineering the form of structural constraints is neither feasible or desirable. Hence, the structural constraints are unknown and must be learned from data automatically. Second, because shape imposes constraints on the relationship between different parts of the object, such as the fact that a person cannot have two heads, it is dependent on the semantics of the image. As a result, the potentials must be capable of representing high-level semantic concepts like ``head'' and would need to have complex non-linear dependence on the input image, which would complicate learning. Finally, because shape simultaneously constrains the labels of many pixels and enforce mutual exclusivity between competing region hypotheses, the potentials would need to be of higher order and non-submodular, often making inference intractable. 

Compared to the traditional single-step structured prediction paradigm, the proposed multi-step prediction procedure is more powerful because it is easier to model local corrections than the global structure. This can be viewed geometrically -- a single-step prediction procedure effectively attempts to model the manifold defined by the structure directly, the geometry of which could be very complex. In contrast, our multi-step procedure learns to model the gradient of an implicit function whose level set defines the manifold, which tends to have much simpler geometry. Because it is possible to recover the manifold, which is a level set of an implicit function, from the gradient of the function, learning the gradient suffices for modelling structure. 

\subsection{Implementation Details}

We modify the architecture introduced by Hariharan et al. \cite{hypercolumn} as follows. Because shape is only expected to be consistent for objects in the same category, we make the weights of the first layer category-dependent by adding twenty channels to the input layer, each corresponding to a different object category. The channel that corresponds to the category given by the detection system contains the heatmap prediction from the previous step, and channels corresponding to other categories are filled with zeros. To prepare the input to the CNN, patches inside the bounding boxes generated by the detection system are extracted and anisotropically scaled to $224 \times 224$ and the ground truth segmentation mask is transformed accordingly. Because the heatmap prediction from the preceding step is $50 \times 50$, we upsample it to $224 \times 224$ using bilinear interpolation before feeding it in as input. To ensure learning is well-conditioned, the heatmap prediction is rescaled and centred element-wise to lie in the range $[-127, 128]$ and the weights corresponding to the additional channels are initialized randomly with the same standard deviation as that of the weights corresponding to the colour channels. 

The training set includes all detection boxes that overlap with the ground truth bounding boxes by more than 70\%. At training time, boxes are uniformly sampled by category, and the weights for upsampled patches are set proportionally to their original areas for the purposes of computing the loss. The weights for all layers that are present in the VGG 16-layer architecture are initialized from the weights finetuned on the detection task and the weights for all other layers are initialized randomly. The loss function is the sum of the pixel-wise negative log likelihoods of the ground truth. The net is trained end-to-end using SGD on mini-batches of 32 patches with a learning rate of $5 \times 10^{-5}$ and momentum of $0.9$. We perform four stages of training and train for 30K, 42.5K, 50K and 20K iterations in stages one, two, three and four respectively. We find that the inference procedure typically converges after three steps and so we use three iterations at test time. 

We can optionally perform post-processing by projecting to superpixels. To generate region predictions from heatmaps, we colour in a pixel or superpixel if the mean heat intensity inside a pixel or superpixel is greater than 40\%. Finally, we can rescore the detections in the same manner as \cite{hypercolumn} by training support vector machines (SVMs) on features computed on the bounding box and the region predictions. To construct the training set, we take all bounding box detections that pass non-maximum suppression (NMS) using a bounding box overlap threshold of $70\%$ and include those that overlap with the ground truth by more than $70\%$ as positive instances and those by less than $50\%$ as negative instances. To compute the features, we feed in the original image patch and the patch with the region background masked out to two CNNs trained as described in \cite{sds}. To obtain the final set of detections, we compute scores using the trained SVMs and apply NMS using a region overlap threshold of $30\%$. 

\subsection{Evaluation}

We evaluate the proposed method in terms of region average precision ($\mathrm{AP}^{r}$), which is introduced by \cite{sds}. Region average precision is defined in the same way as the standard average precision metric used for the detection task, with the difference being the computation of overlap between the prediction and the ground truth. For instance segmentation, overlap is defined as the pixel-wise intersection-over-union (IoU) of the region prediction and the ground truth segmentation mask, instead of the IoU of their respective bounding boxes. We evaluate against the SBD instance segmentation annotations on the PASCAL VOC 2012 validation set. 

\section{Experiments}

\begin{figure}[h]
    \centering
    \includegraphics[width=1.0\columnwidth]{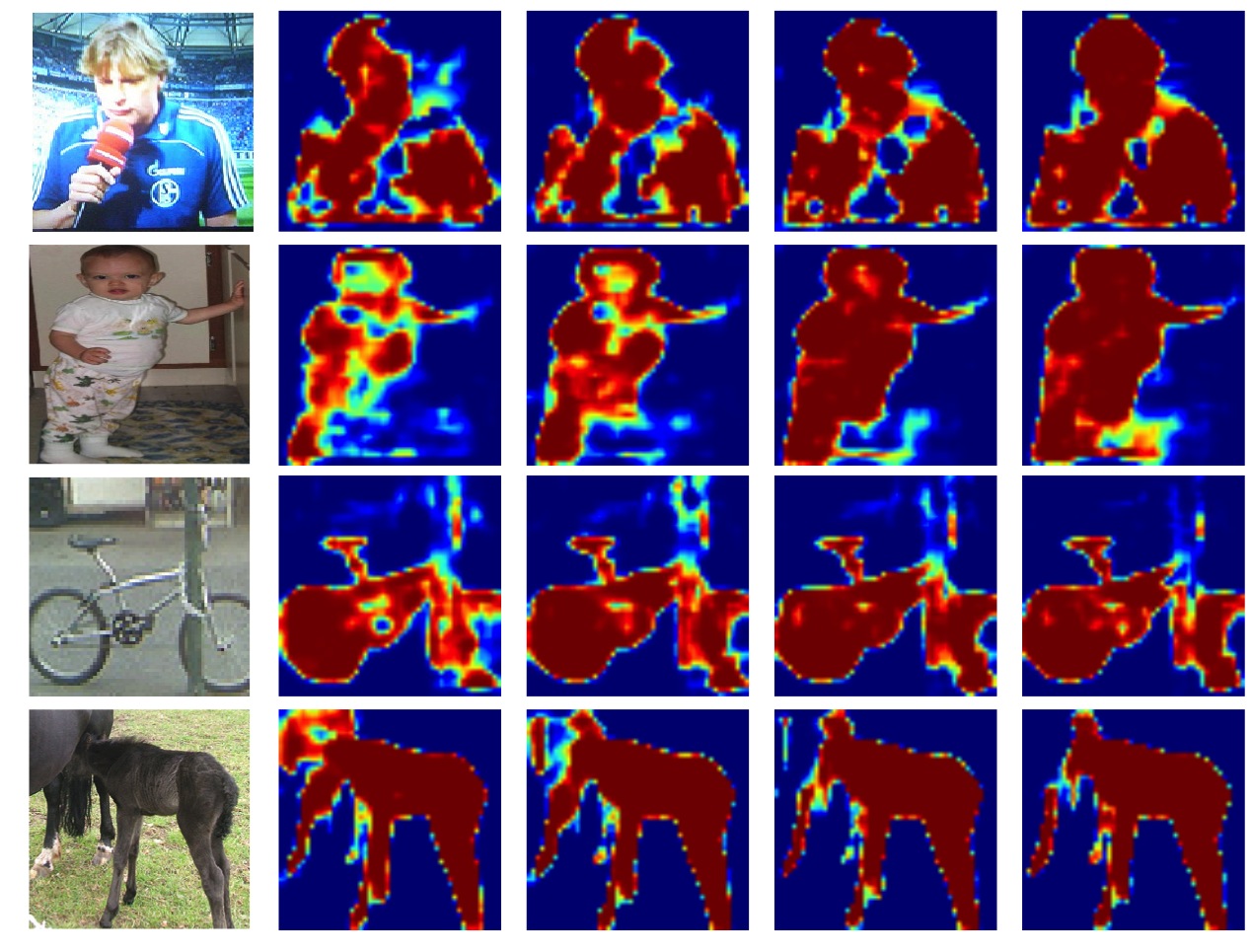}
    \caption{Heatmap predictions on images from the PASCAL VOC 2012 validation set after each stage of training. Best viewed in colour. }
    \label{fig:ief_training}
\end{figure}

First, we visualize the improvement in prediction accuracy as training progresses. In Figure \ref{fig:ief_training}, we show the pixel-wise heatmap predictions on image patches from the PASCAL VOC 2012 validation set after each stage of training. As shown, prediction quality steadily improves with each successive stage of training. Initially, the model is only able to identify some parts of the object; with each stage of training, it learns to recover additional parts of the object that were previously missed. After four stages of training, the model is able to correctly identify most parts belonging to the object. This indicates that the model is able to learn to make local corrections to its predictions in each stage. After four stages of training, the predictions are reasonably visually coherent and consistent with the underlying structure of the output space. Interestingly, the model gradually learns to suppress parts of other objects, as shown by the predictions on the bicycle and horse images, where the model learns to suppress parts of the pole and the other horse in later stages. 

\begin{figure*}[t]
    \centering
    \includegraphics[width=1.0\textwidth]{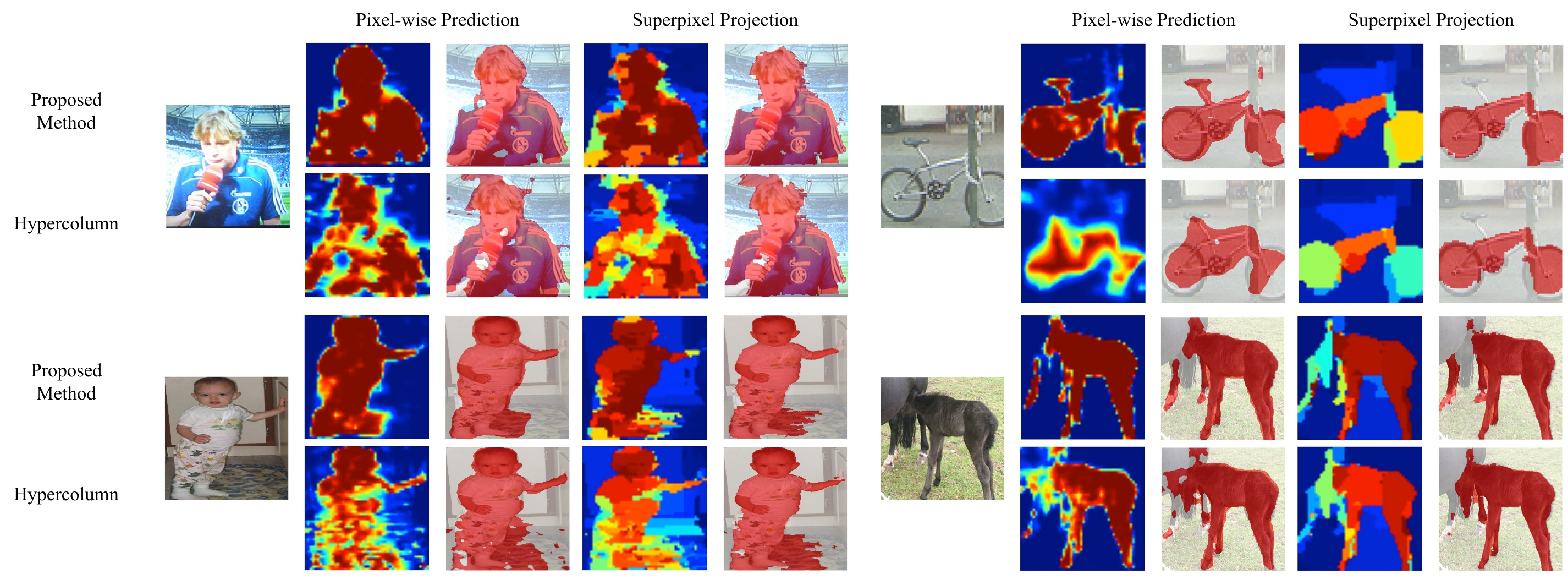}
    \caption{Comparison of heatmap and region predictions produced by the proposed method and the vanilla hypercolumn net on images from the PASCAL VOC 2012 validation set. Best viewed in colour. }
    \label{fig:preds}
\end{figure*}

Next, we compare the performance of the proposed method with that of existing methods. As shown in Table \ref{tab:top_line_map_comp}, the proposed method outperforms all existing methods in terms of mean $\mathrm{AP}^{r}$ at both 50\% and 70\%. We analyze performance at a more granular level by comparing the proposed method to the state-of-the-art method, the hypercolumn net \cite{hypercolumn}, under three settings: without superpixel projection, with superpixel projection and with superpixel projection and rescoring. As shown in Table \ref{tab:map_comp}, the proposed method achieves higher mean $\mathrm{AP}^{r}$ at 50\% and 70\% than the state-of-the-art in each setting. In particular, the proposed method achieves an $9.3$-point gain over the state-of-the-art in terms of its raw pixel-wise prediction performance at $70\%$ overlap. This indicates the raw heatmaps produced by the proposed method are more accurate than those produced by the vanilla hypercolumn net. As a result, the proposed method requires less reliance on post-processing. We confirm this intuition by visualizing the heatmaps in Figure \ref{fig:preds}. When superpixel projection is applied, the proposed method improves performance by $1.7$ points and $3.8$ points at $50\%$ and $70\%$ overlaps respectively. With rescoring, the proposed method obtains a mean $\mathrm{AP}^{r}$ of $63.6\%$ at $50\%$ overlap and $43.3\%$ at $70\%$ overlap, which represent the best performance on the instance segmentation task to date. We break down performance by category under each setting in the supplementary material.  

\begin{table}
\centering
\footnotesize
\begin{tabular}{l c c}
\toprule 
Method & $\mathrm{mAP}^{r}$ at 50\% & $\mathrm{mAP}^{r}$ at 70\%\\
\midrule 
O$_2$P \cite{CarreiraECCV2012} & $25.2$ & $-$ \\ 
SDS \cite{sds} & $49.7$ & $25.3$ \\
CFM \cite{DaiCVPR2015} & $60.7$ & $39.6$ \\
Hypercolumn \cite{hypercolumn} & $62.4$ & $39.4$ \\
Proposed Method & $\mathbf{63.6}$ & $\mathbf{43.3}$ \\
\bottomrule
\end{tabular}
\caption{Performance of the proposed method compared to existing methods. }
\label{tab:top_line_map_comp}
\end{table}

\begin{table}
\centering
\footnotesize
\begin{tabular}{l c c}
\toprule 
Method and Setting & $\mathrm{mAP}^{r}$ at 50\% & $\mathrm{mAP}^{r}$ at 70\%\\
\midrule
\emph{Raw pixel-wise prediction:} \\
\; Hypercolumn \cite{hypercolumn} & $56.1$ & $29.4$ \\
\; Proposed Method & $\mathbf{60.1}$ & $\mathbf{38.7}$ \\
\midrule
\emph{With superpixel projection:} \\
\; Hypercolumn \cite{hypercolumn} & $58.6$ & $36.4$ \\
\; Proposed Method & $\mathbf{60.3}$ & $\mathbf{40.2}$ \\
\midrule
\emph{With superpixel projection} \\
\emph{and rescoring:} \\
\; Hypercolumn \cite{hypercolumn} & $62.4$ & $39.4$ \\
\; Proposed Method & $\mathbf{63.6}$ & $\mathbf{43.3}$ \\
\bottomrule
\end{tabular}
\caption{Performance comparison of the proposed method and the state-of-the-art under different settings. }
\label{tab:map_comp}
\end{table}

We examine heatmap and region predictions of the proposed method and the vanilla hypercolumn net, both with and without applying superpixel projection. As shown in Figure \ref{fig:preds}, the pixel-wise heatmap predictions produced by the proposed method are generally more visually coherent than those produced by the vanilla hypercolumn net. In particular, the proposed method predicts regions that are more consistent with shape. For example, the heatmap predictions produced by the proposed method for the sportscaster and the toddler images contain less noise and correctly identify most foreground pixels with high confidence. In contrast, the heatmap predictions produced by the hypercolumn net are both noisy and inconsistent with the typical shape of persons. On the bicycle image, the proposed method is able to produce a fairly accurate segmentation, whereas the hypercolumn net  largely fails to find the contours of the bicycle. On the horse image, the proposed method correctly identifies the body and the legs of the horse. It also incorrectly hallucinates the head of the horse, which is actually occluded; this mistake is reasonable given the similar appearance of adjacent horses. This effect provides some evidence that the method is able to learn a shape prior successfully; because the shape prior discounts the probability of seeing a headless horse, it causes the model to hallucinate a head. On the other hand, the hypercolumn net chooses to hedge its bets on the possible locations of the head and so the resulting region prediction is noisy in the area near the expected location of the head. Notably, the region predictions generated by the proposed method also tend to contain fewer holes and have smoother contours than those produced by the hypercolumn net, which is apparent in the case of the sportscaster and toddler images. This suggests that the model is able to learn a prior favouring the contiguity of regions and smoothness of region contours. More examples of heatmap and region predictions can be found in the supplementary material. 

Applying superpixel projection significantly improves the region predictions of the vanilla hypercolumn net. It effectively smoothes out noise in the raw heatmap predictions by averaging the heat intensities over all pixels in a superpixel. As a result, the region predictions contain fewer holes after applying superpixel projection, as shown by the predictions on the sportscaster and toddler images. Superpixel projection also ensures that the region predictions conform to the edge contours in the image, which can result in a significant improvement if the raw pixel-wise region prediction is very poor, as is the case on the bicycle image. On the other hand, because the raw pixel-wise predictions of the proposed method are generally less noisy and have more accurate contours than those of the hypercolumn net, superpixel projection does not improve the quality of predictions as significantly. In some cases, it may lead to a performance drop, as pixel-wise prediction may capture details that are missed by the superpixel segmentation. As an example, on the bicycle image, the seat is originally segmented correctly in the pixel-wise prediction, but is completely missed after applying superpixel projection. Therefore, superpixel projection has the effect of masking prediction errors and limits performance when the quality of pixel-wise predictions becomes better than that of the superpixel segmentation. 

\begin{figure}[h]
    \centering
    \includegraphics[width=1.0\columnwidth]{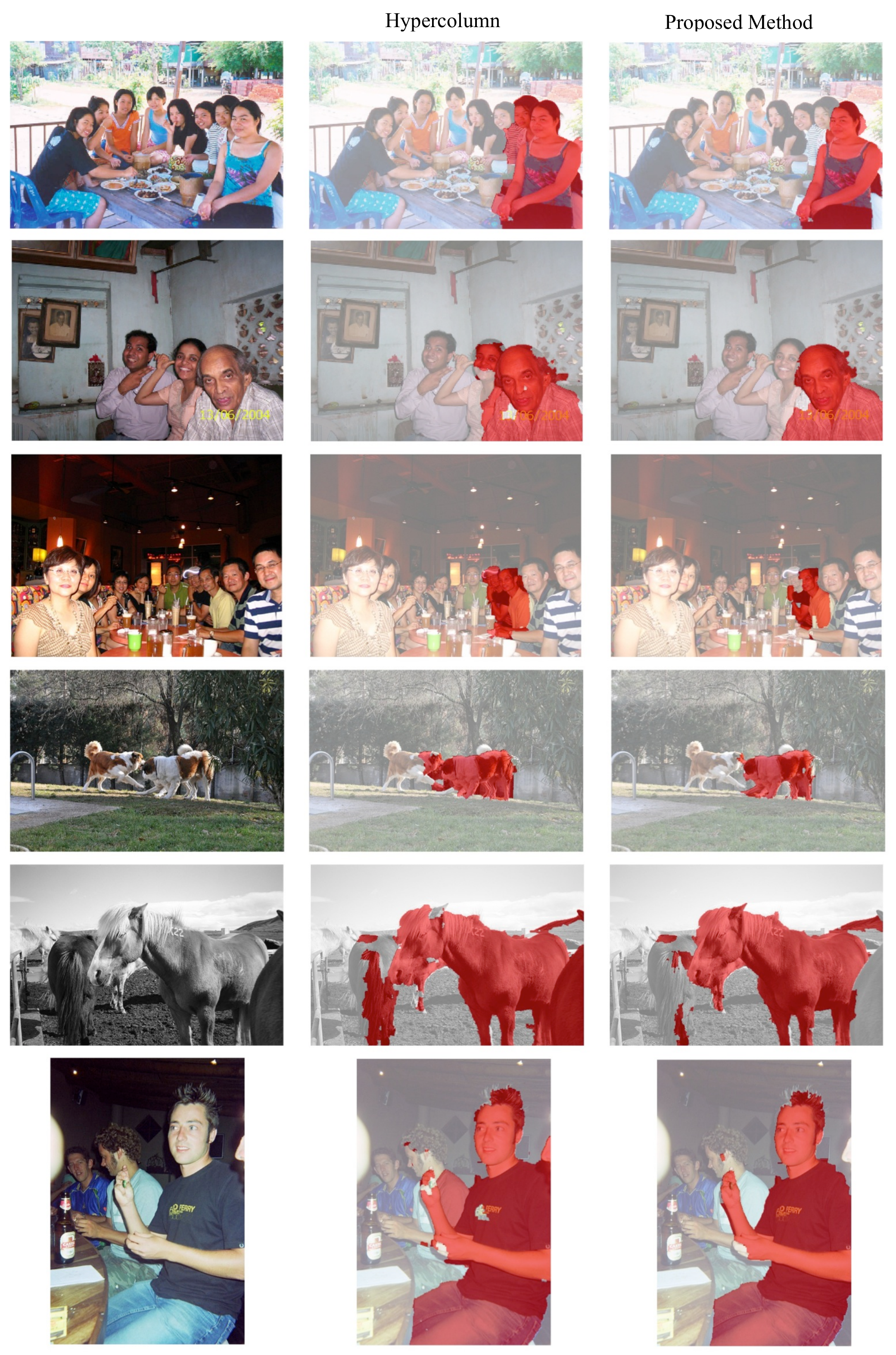}
    \caption{Region predictions on images with challenging scene configurations. }
    \label{fig:challenging_img}
\end{figure}

We find that the proposed method is able to avoid some of the mistakes made by the vanilla hypercolumn net on images with challenging scene configurations, such as those depicting groups of people or animals. On such images, the hypercolumn net sometimes includes parts of adjacent persons in region predictions. Several examples are shown in Figure \ref{fig:challenging_img}, in which region predictions contain parts from different people or animals. The proposed method is able to suppress parts of adjacent objects and correctly exclude them from region predictions, suggesting that the learned shape prior is able to help the model disambiguate region hypotheses that are otherwise consistent with local appearance cues. 

\begin{figure}[h]
    \centering
    \includegraphics[width=1.0\columnwidth]{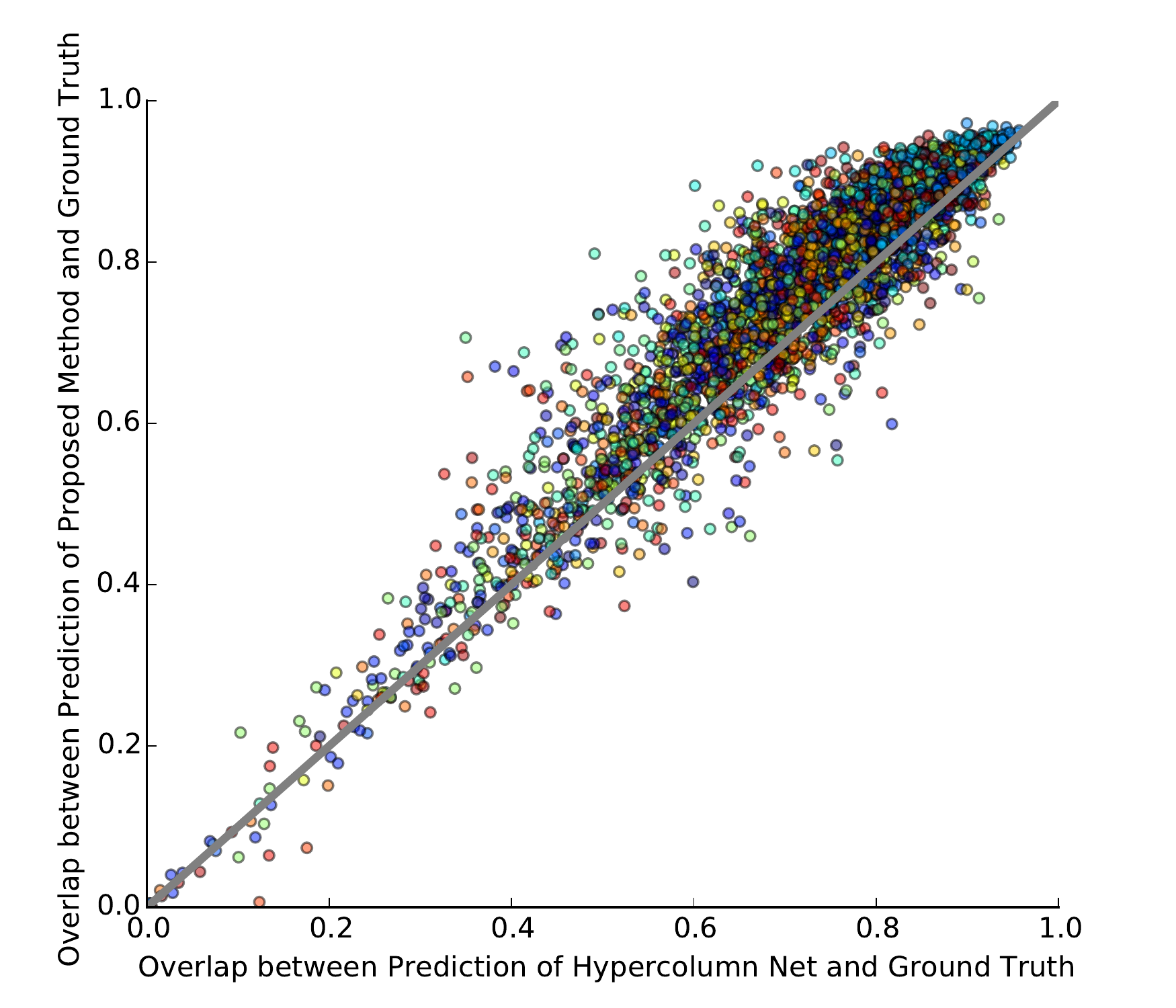}
    \caption{Comparison of maximum overlap of region predictions produced by the vanilla hypercolumn net and the proposed method with the ground truth. Each data point corresponds to a bounding box detection and the colour of each data point denotes the category of the detection. Points that lie above the diagonal represent detections for which the region predictions produced by the proposed method are more accurate than those produced by the hypercolumn net. }
    \label{fig:overlap_comparison}
\end{figure}

We now analyze the improvement in overlap between region predictions and the ground truth segmentation masks at the level of individual detections. In Figure \ref{fig:overlap_comparison}, we plot the maximum overlap of the pixel-wise region prediction produced by the proposed method with the ground truth against that of the region prediction generated by the vanilla hypercolumn net for each of the top 200 detections in each category. So, in this plot, any data point above the diagonal represents a detection for which the proposed method produces a more accurate region prediction than the hypercolumn net. We find overlap with ground truth improves for 76\% of the detections, degrades for 15.6\% of the detections and remains the same for the rest. This is reflected in the plot, where the vast majority of data points lie above the diagonal, indicating that the proposed method improves the accuracy of region predictions for most detections. 

Remarkably, for detections on which reasonably good overlap is achieved using the vanilla hypercolumn net, which tend to correspond to bounding boxes that are well-localized, the proposed method can improve overlap by $~15\%$ in many cases. Furthermore, the increase in overlap tends to be the greatest for detections on which the hypercolumn net achieves $~75\%$ overlap; when the proposed method is used, overlap for these detections at times reach more than $90\%$. This is particularly surprising given that improving upon good predictions is typically challenging. Such a performance gain is conceptually difficult to achieve without leveraging structure in the output. This suggests that the proposed method is able to use the priors it learned to further refine region predictions that are already very accurate. 

Finally, we conduct an experiment to test whether the proposed method is indeed able to learn a shape prior more directly. To this end, we select an image patch from the PASCAL VOC 2012 validation set that contains little visually distinctive features, so that it does not resemble an object from any of the categories. We then feed the patch into the model along with an arbitrary category label, which essentially forces the model to try to interpret the image as that of an object of the particular category. We are interested in examining if the model is able to hallucinate a region that is both consistent with the input image and resembles an object from the specified category. 

\begin{figure}[h]
    \centering
    \includegraphics[width=1.0\columnwidth]{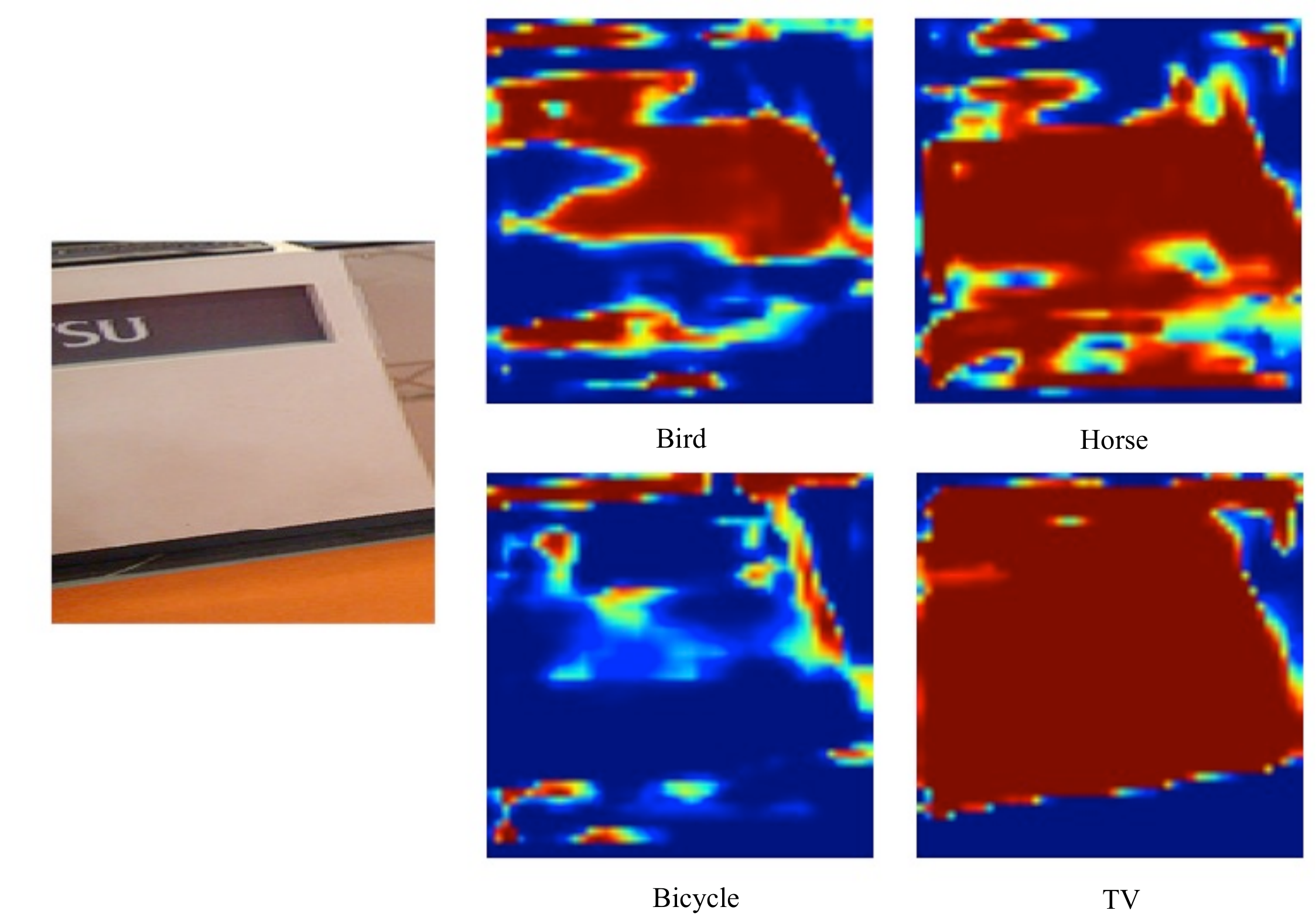}
    \caption{Heatmap predictions of the proposed method under different settings of category. As shown, the model is able to hallucinate plausible shapes that correspond to the specified categories. }
    \label{fig:hallucinations}
\end{figure}

Figure \ref{fig:hallucinations} shows the input image and the resulting heatmap predictions under different settings of category. As shown, when the category is set to bird, the heatmap prediction resembles the body and the wing of a bird. When the category is set to horse, the model hallucinates the body and the legs of a horse. Interestingly, the wing of the bird and the legs of the horse are hallucinated even though there are no corresponding contours that resemble these parts in the input image. When the category is set to bicycle, the model interprets the edges in the input image as the frame of a bicycle, which contrasts with the heatmap prediction when the category is set to television, which is not sensitive to thin edges in the input image and instead contains a large contiguous box that resembles the shape of a television set. 

\section{Conclusion}

We presented a method that is able to take advantage of the implicit structure that underlies the output space when making predictions. The method does not require manual specification of the form of the structure a priori and is able to discover salient structure from the data automatically. We applied the method to the instance segmentation task and showed that the method automatically learns a prior on shape, contiguity of regions and smoothness of region contours. We also demonstrated state-of-the-art performance using the method, which achieves a mean $\mathrm{AP}^{r}$ of $63.6\%$ and $43.3\%$ at $50\%$ and $70\%$ overlaps respectively. The method is generally applicable to all tasks that require the prediction of a pixel-wise labelling of the input image; we hope the success we demonstrated on instance segmentation will encourage application to other such tasks and further exploration of the method. 

\paragraph{Acknowledgements.} This work was supported by ONR MURI N00014-09-1-1051 and ONR MURI N00014-14-1-0671. Ke Li thanks the Natural Sciences and Engineering Research Council of Canada (NSERC) for fellowship support. The authors also thank NVIDIA Corporation for the donation of GPUs used for this research. 

{\small
\bibliographystyle{ieee}
\bibliography{iterative_inst_seg}
}

\newpage

\onecolumn

\title{Iterative Instance Segmentation \\ \vspace{5pt} \large Supplementary Material}

\author{Ke Li\\
UC Berkeley\\
{\tt\small ke.li@eecs.berkeley.edu}
\and
Bharath Hariharan\\
Facebook AI Research\\
{\tt\small bharathh@fb.com}
\and
Jitendra Malik\\
UC Berkeley\\
{\tt\small malik@eecs.berkeley.edu}
}

\maketitle
\thispagestyle{empty}

\section{Per-Category Performance Comparison}

We report the per-category performance of the proposed method compared to the state-of-the-art below. 

\begin{table}[h]
\centering
\footnotesize
\begin{tabular}{lccccccccccc}
\toprule 
Method and Setting & aero & bike & bird & boat & bottle & bus & car & cat & chair & cow\\
\midrule
\emph{Raw pixel-wise prediction:} \\
\; Hypercolumn [16] & $74.8$ & $57.4$ & $61.6$ & $38.3$ & $32.3$ & $79.1$ & $57.9$ & $82.3$ & $20.8$ & $55.2$ \\
\; Proposed Method & $\mathbf{77.3}$ & $\mathbf{65.3}$ & $\mathbf{65.5}$ & $\mathbf{42.5}$ & $\mathbf{35.4}$ & $\mathbf{80.3}$ & $\mathbf{62.2}$ & $\mathbf{83.9}$ & $\mathbf{27.2}$ & $\mathbf{61.6}$ \\
\midrule
\emph{With superpixel projection:} \\
\; Hypercolumn [16] & $\mathbf{76.4}$ & $63.4$ & $63.8$ & $\mathbf{42.9}$ & $32.3$ & $80.0$ & $59.5$ & $\mathbf{82.4}$ & $27.5$ & $59.9$ \\
\; Proposed Method & $76.3$ & $\mathbf{64.9}$ & $\mathbf{65.1}$ & $42.6$ & $\mathbf{35.1}$ & $\mathbf{80.6}$ & $\mathbf{61.2}$ & $80.9$ & $\mathbf{28.3}$ & $\mathbf{61.7}$ \\
\midrule
\emph{With superpixel projection and rescoring:} \\
\; Hypercolumn [16] & $78.2$ & $67.0$ & $68.2$ & $46.9$ & $42.0$ & $\mathbf{82.9}$ & $66.7$ & $\mathbf{85.0}$ & $\mathbf{31.2}$ & $\mathbf{66.7}$ \\
\; Proposed Method & $\mathbf{79.2}$ & $\mathbf{67.9}$ & $\mathbf{70.0}$ & $\mathbf{47.9}$ & $\mathbf{45.3}$ & $81.6$ & $\mathbf{68.8}$ & $84.1$ & $30.4$ & $65.5$ \\
\midrule
Method and Setting & table & dog & horse & mbike & person & plant & sheep & sofa & train & tv & $\mathrm{mAP}^{r}$\\
 \midrule
\emph{Raw pixel-wise prediction:} \\
\; Hypercolumn [16] & $27.5$ & $80.0$ & $65.3$ & $69.6$ & $52.4$ & $27.5$ & $58.1$ & $44.7$ & $77.5$ & $59.9$ & $56.1$ \\
\; Proposed Method & $\mathbf{32.4}$ & $\mathbf{82.3}$ & $\mathbf{70.9}$ & $\mathbf{71.4}$ & $\mathbf{63.1}$ & $\mathbf{31.3}$ & $\mathbf{63.6}$ & $\mathbf{44.9}$ & $\mathbf{78.3}$ & $\mathbf{62.4}$ & $\mathbf{60.1}$ \\
\midrule
\emph{With superpixel projection:} \\
\; Hypercolumn [16] & $30.1$ & $81.0$ & $69.3$ & $70.6$ & $60.8$ & $27.3$ & $60.7$ & $45.6$ & $77.3$ & $\mathbf{61.8}$ & $58.6$ \\
\; Proposed Method & $\mathbf{33.6}$ & $\mathbf{82.2}$ & $\mathbf{71.2}$ & $\mathbf{71.9}$ & $\mathbf{63.7}$ & $\mathbf{31.1}$ & $\mathbf{65.1}$ & $\mathbf{49.6}$ & $\mathbf{78.9}$ & $61.5$ & $\mathbf{60.3}$ \\
\midrule
\emph{With superpixel projection and rescoring:} \\
\; Hypercolumn [16] & $30.1$ & $82.0$ & $73.1$ & $73.3$ & $64.6$ & $37.3$ & $68.9$ & $41.4$ & $75.3$ & $67.9$ & $62.4$ \\
\; Proposed Method & $\mathbf{31.8}$ & $\mathbf{83.6}$ & $\mathbf{75.5}$ & $\mathbf{74.5}$ & $\mathbf{66.6}$ & $\mathbf{37.7}$ & $\mathbf{70.6}$ & $\mathbf{44.7}$ & $\mathbf{77.7}$ & $\mathbf{68.7}$ & $\mathbf{63.6}$ \\
\bottomrule
\end{tabular}
\caption{Per-category $\mathrm{AP}^{r}$ at 50\% overlap achieved by the proposed method compared to the state-of-the-art on the PASCAL VOC 2012 validation set. }
\label{tab:categ_specific_res_50}
\end{table}
\begin{table}[h]
\centering
\footnotesize
\begin{tabular}{lccccccccccc}
\toprule 
Setting & aero & bike & bird & boat & bottle & bus & car & cat & chair & cow\\
\midrule
\emph{Raw pixel-wise prediction:} \\
\; Hypercolumn [16] & $52.4$ & $18.6$ & $23.2$ & $15.1$ & $17.3$ & $68.0$ & $36.5$ & $53.5$ & $2.1$ & $26.9$ \\
\; Proposed Method & $\mathbf{61.8}$ & $\mathbf{31.5}$ & $\mathbf{42.0}$ & $\mathbf{22.0}$ & $\mathbf{22.7}$ & $\mathbf{72.4}$ & $\mathbf{44.8}$ & $\mathbf{65.4}$ & $\mathbf{7.2}$ & $\mathbf{37.6}$ \\
\midrule
\emph{With superpixel projection:} \\
\; Hypercolumn [16] & $53.3$ & $26.4$ & $35.4$ & $\mathbf{24.0}$ & $22.6$ & $71.0$ & $41.8$ & $61.4$ & $8.4$ & $36.0$ \\
\; Proposed Method & $\mathbf{57.4}$ & $\mathbf{33.2}$ & $\mathbf{42.9}$ & $23.1$ & $\mathbf{23.4}$ & $71.0$ & $\mathbf{44.9}$ & $\mathbf{64.4}$ & $\mathbf{10.8}$ & $\mathbf{40.6}$ \\
\midrule
\emph{With superpixel projection and rescoring:} \\
\; Hypercolumn [16] & $55.6$ & $28.7$ & $41.2$ & $\mathbf{26.8}$ & $25.5$ & $73.5$ & $45.2$ & $64.7$ & $10.6$ & $42.3$ \\
\; Proposed Method & $\mathbf{61.9}$ & $\mathbf{35.1}$ & $\mathbf{44.4}$ & $26.4$ & $\mathbf{29.6}$ & $\mathbf{74.0}$ & $\mathbf{48.7}$ & $\mathbf{66.8}$ & $\mathbf{10.9}$ & $\mathbf{48.4}$ \\
\midrule
Setting & table & dog & horse & mbike & person & plant & sheep & sofa & train & tv & $\mathrm{mAP}^{r}$\\
 \midrule
\emph{Raw pixel-wise prediction:} \\
\; Hypercolumn [16] & $8.1$ & $47.4$ & $20.7$ & $35.4$ & $15.6$ & $7.2$ & $28.4$ & $14.9$ & $53.2$ & $44.3$ & $29.4$ \\
\; Proposed Method & $\mathbf{10.4}$ & $\mathbf{60.4}$ & $\mathbf{39.6}$ & $\mathbf{41.9}$ & $\mathbf{32.5}$ & $\mathbf{12.0}$ & $\mathbf{40.9}$ & $\mathbf{19.9}$ & $\mathbf{58.8}$ & $\mathbf{50.8}$ & $\mathbf{38.7}$ \\
\midrule
\emph{With superpixel projection:} \\
\; Hypercolumn [16] & $10.9$ & $58.1$ & $32.8$ & $41.2$ & $27.6$ & $10.2$ & $37.6$ & $25.6$ & $56.4$ & $48.3$ & $36.4$ \\
\; Proposed Method & $\mathbf{14.3}$ & $\mathbf{62.7}$ & $\mathbf{42.1}$ & $\mathbf{44.1}$ & $\mathbf{36.2}$ & $\mathbf{11.6}$ & $\mathbf{44.4}$ & $\mathbf{27.6}$ & $\mathbf{60.1}$ & $\mathbf{49.7}$ & $\mathbf{40.2}$ \\
\midrule
\emph{With superpixel projection and rescoring:} \\
\; Hypercolumn [16] & $12.3$ & $60.8$ & $41.7$ & $42.1$ & $27.3$ & $15.5$ & $45.2$ & $\mathbf{23.9}$ & $56.6$ & $47.8$ & $39.4$ \\
\; Proposed Method & $\mathbf{13.6}$ & $\mathbf{64.0}$ & $\mathbf{53.0}$ & $\mathbf{46.8}$ & $\mathbf{33.0}$ & $\mathbf{19.0}$ & $\mathbf{51.0}$ & $23.7$ & $\mathbf{62.2}$ & $\mathbf{53.9}$ & $\mathbf{43.3}$ \\
\bottomrule
\end{tabular}
\caption{Per-category $\mathrm{AP}^{r}$ at 70\% overlap achieved by the proposed method compared to the state-of-the-art on the PASCAL VOC 2012 validation set. }
\label{tab:categ_specific_res_70}
\end{table}

\newpage
\section{Additional Visualizations}

The following are predictions of the proposed method and the vanilla hypercolumn net on additional images from various categories. 

\begin{figure}[h]
    \centering
    \includegraphics[width=1.0\textwidth]{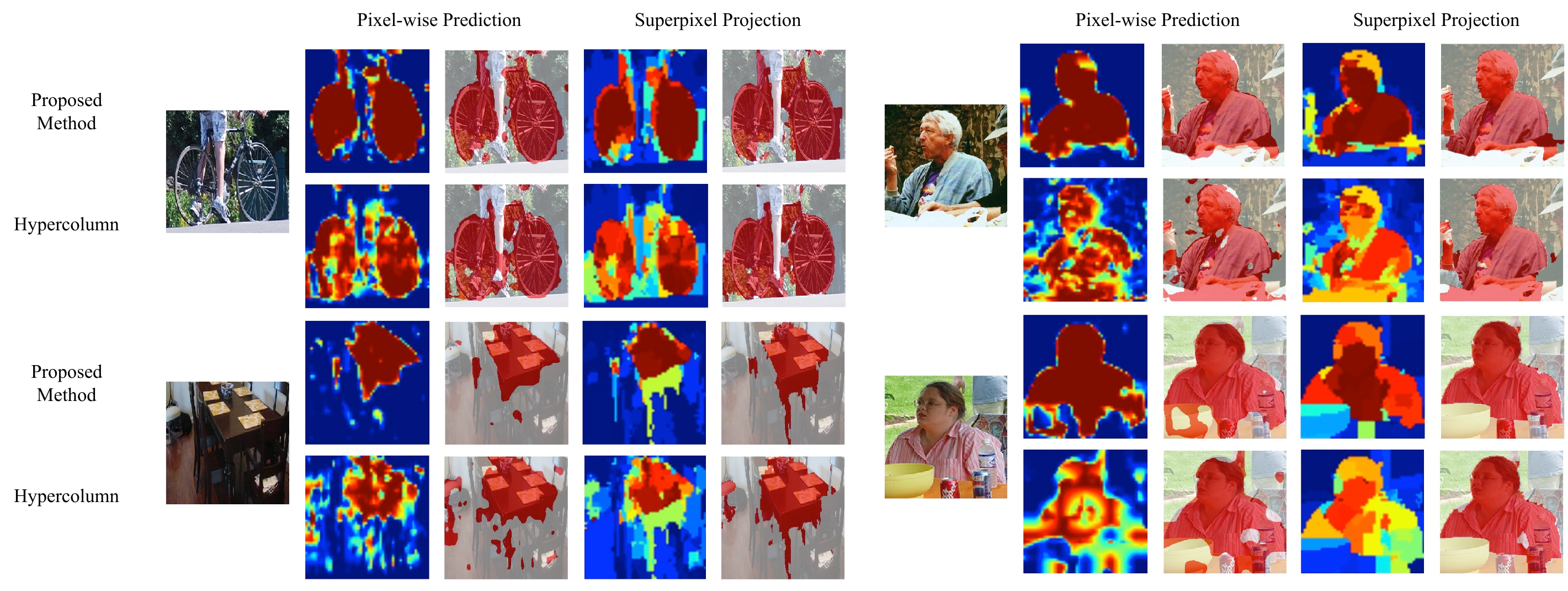}
    \caption{Comparison of heatmap and region predictions produced by the proposed method and the vanilla hypercolumn net on images from the PASCAL VOC 2012 validation set. Best viewed in colour. }
    \label{fig:supp3}
\end{figure}

\begin{figure}[h]
    \centering
    \includegraphics[width=1.0\textwidth]{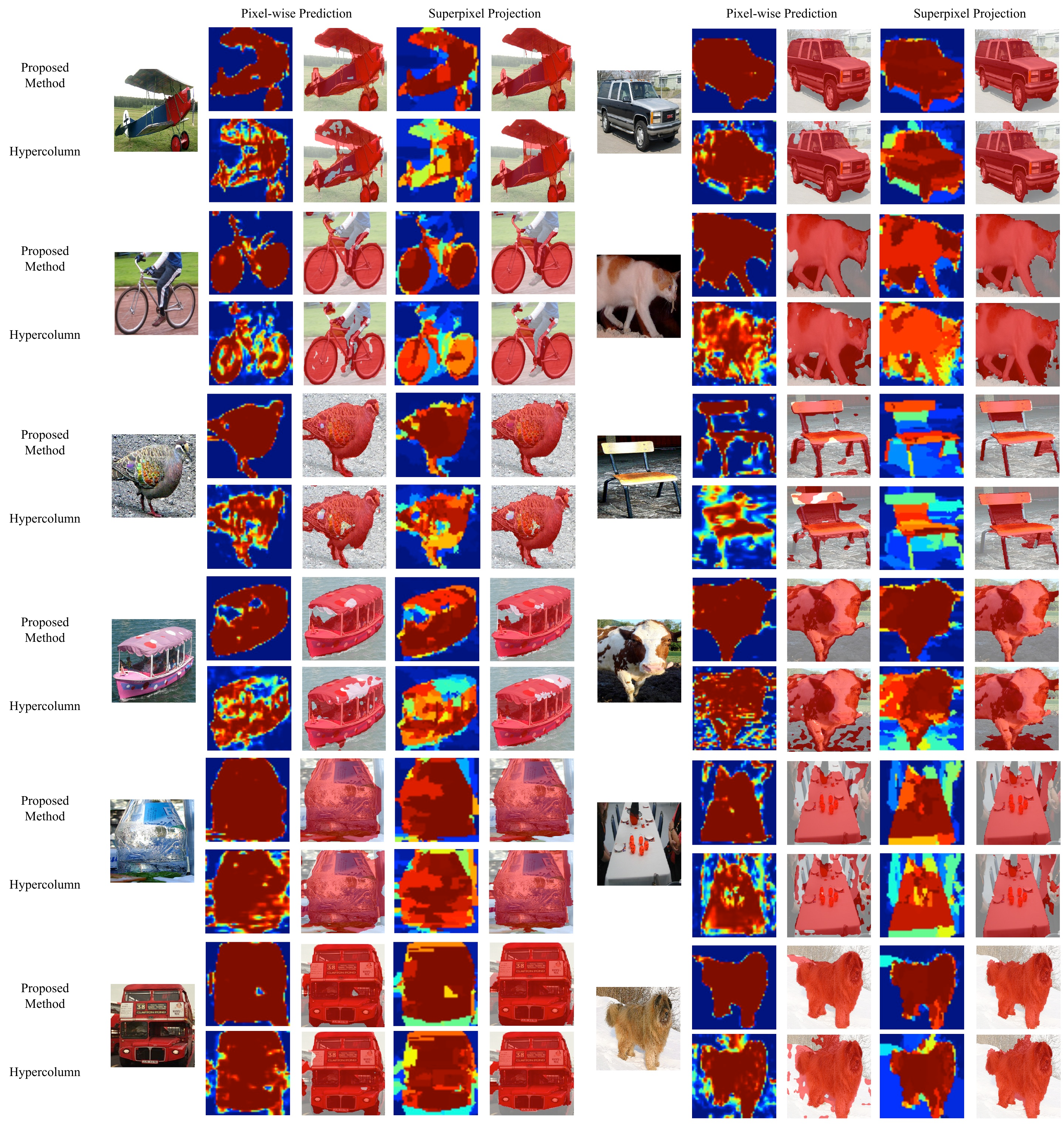}
    \caption{Comparison of heatmap and region predictions produced by the proposed method and the vanilla hypercolumn net on images from the PASCAL VOC 2012 validation set. Best viewed in colour. }
    \label{fig:supp1}
\end{figure}

\begin{figure}[h]
    \centering
    \includegraphics[width=1.0\textwidth]{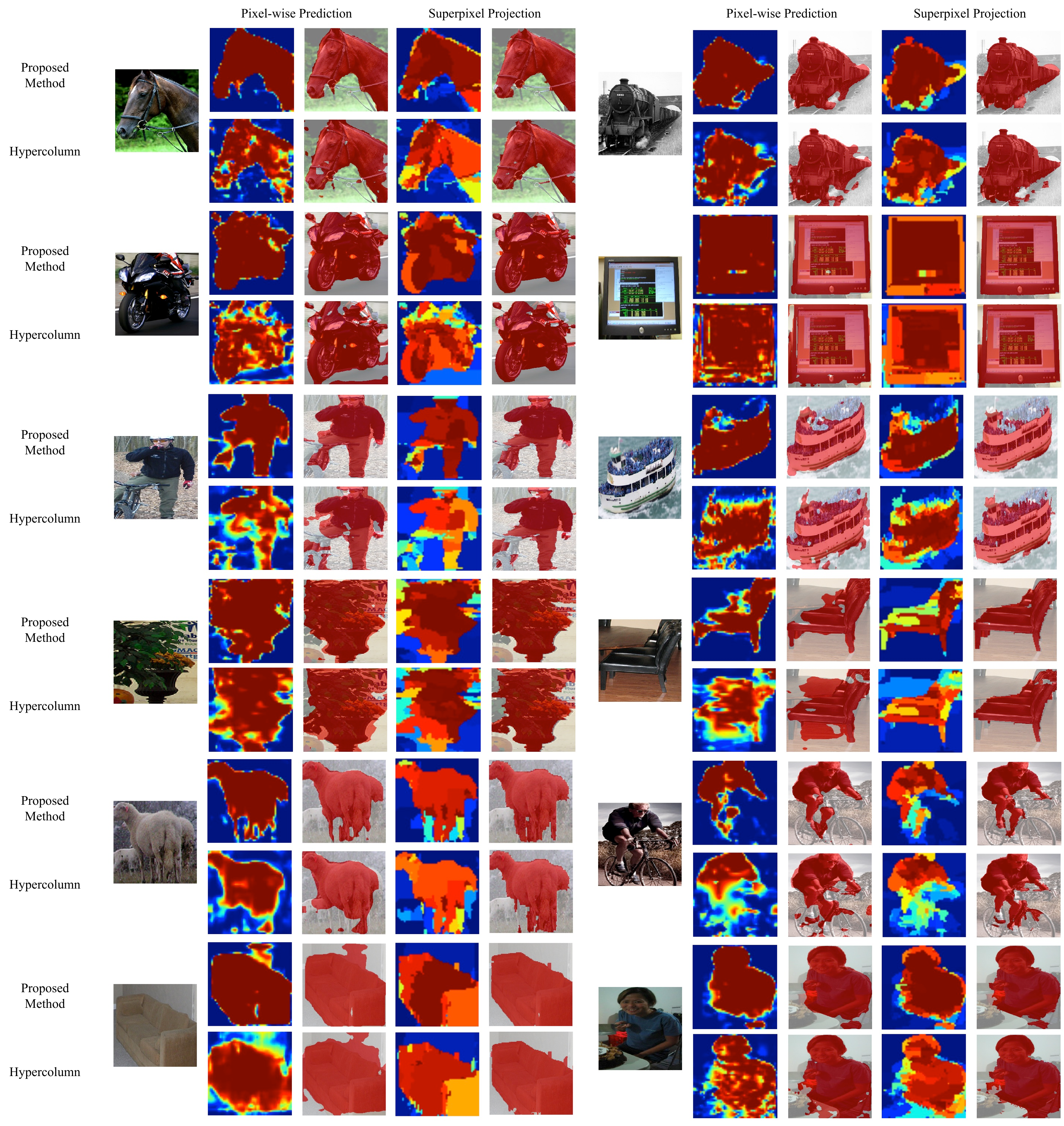}
    \caption{Comparison of heatmap and region predictions produced by the proposed method and the vanilla hypercolumn net on images from the PASCAL VOC 2012 validation set. Best viewed in colour. }
    \label{fig:supp2}
\end{figure}

\end{document}